%% file: emnlp2020.tex
\documentclass[11pt,a4paper]{article}
\usepackage[hyperref]{emnlp2020}
\usepackage{times}
\usepackage{latexsym}
\usepackage{multirow}

\usepackage{amsmath}
\usepackage{amssymb}
\usepackage{graphicx}
\usepackage{subfigure}
\usepackage{booktabs}
\usepackage{soul}
\usepackage{color}
\usepackage{xspace}

\newcommand{\pink}[1]{\textcolor[cmy]{0,1,0}{#1}}

\usepackage{enumitem}

\usepackage{microtype}
\usepackage{enumitem}
\setenumerate[1]{itemsep=0pt,partopsep=0pt,parsep=\parskip,topsep=1pt}
\setitemize[1]{itemsep=0pt,partopsep=0pt,parsep=\parskip,topsep=1pt}
\setdescription{itemsep=0pt,partopsep=0pt,parsep=\parskip,topsep=5pt}

\aclfinalcopy % Uncomment this line for the final submission
\def\aclpaperid{1066} %  Enter the acl Paper ID here

\newcommand{\hide}[1]{}
\newcommand{\method}{TSMH\xspace}

\title{Language Generation via Combinatorial Constraint Satisfaction: \\
A Tree Search Enhanced Monte-Carlo Approach}

\author{Maosen Zhang$^\dagger$, Nan Jiang$^\dagger$, Lei Li$^\ddagger$, and Yexiang Xue$^\dagger$ \\
$^\dagger$Department of Computer Science, Purdue University, Indiana, USA \\
$^\ddagger$ByteDance AI Lab\\
  \texttt{\{maosen,jiang631,yexiang\}@purdue.edu, lileilab@bytedance.com} \\
  }

\date{}

\begin{document}
\maketitle
\begin{abstract}
\input{0-abstract.tex}
\end{abstract}

\input{1-intro.tex}

\input{2-constr.tex}
\input{3-sample.tex}
\input{4-search.tex}

\input{5-exp.tex}
\input{6-conclude.tex}

\section*{Acknowledgements}
This research was supported by the National Science Foundation (Award number IIS-1850243 and CCF-1918327). The computing infrastructure was partially supported by the Microsoft AI for Earth computing award. The authors would like to thank Mr. Ning Miao for valuable suggestions.

\bibliographystyle{acl_natbib}
\bibliography{emnlp2020}

\clearpage
\input{7-supp}

\end{document}

%% file: 0-abstract.tex
Generating natural language under complex constraints is a principled formulation towards controllable text generation. 
We present a framework to allow specification of combinatorial constraints for sentence generation.
We propose \method \footnote{https://github.com/Milozms/TSMH}, an efficient method to generate high likelihood sentences with respect to a pre-trained language model while satisfying the constraints.
\hide{
We present a constraint satisfaction driven approach for language generation. 
Our approach samples sentences that attain high likelihoods from a pre-trained language model measuring quality and satisfy task-specific constraints.}
% Our approach samples sentences that attain high likelihoods from a pre-trained language model measuring quality while satisfying task-specific combinatorial constraints.
% Our approach samples sentences from a probabilistic distribution proportional to the likelihood of a general-purpose language model, while enforcing that the sentences satisfy task-specific constraints.
% Our approach samples sentences from a probabilistic distribution proportional to the likelihood of a general-purpose language model, while satisfying task-specific constraints.
%
Our approach is highly flexible,  requires no task-specific training, and leverages
efficient constraint satisfaction solving techniques. 
To better handle the combinatorial constraints, a tree search algorithm is embedded into the proposal process of the Markov chain Monte Carlo (MCMC) to explore candidates that satisfy more constraints.
Compared to existing MCMC approaches, our sampling approach has a better mixing performance. 
Experiments show that \method achieves consistent and significant improvement on multiple language generation tasks. 
\hide{
The sentences generated satisfy grammatical constraints better and are more realistic both qualitatively and quantitatively. 
}

% We present a constraint satisfaction driven approach for language generation. Our approach samples sentences that attain high likelihoods from a pre-trained language model measuring quality and satisfy task-specific constraints. Our approach is highly flexible,  requires no task-specific training, and leverages efficient constraint satisfaction tools. To better handle the combinatorial constraints, a tree search algorithm is embedded into the proposal process of the Markov Chain Monte Carlo (MCMC) for our task to suggest candidate proposals that satisfy more constraints. Compared to MCMC, our sampling approach has a better mixing performance. The sentences generated satisfy grammatical constraints better and are more realistic both qualitatively and quantitatively. 

%% file: 1-intro.tex
\section{Introduction}
% \vspace{-0.2cm}
\begin{figure}[t]
    \centering
    \includegraphics[width=.95\linewidth]{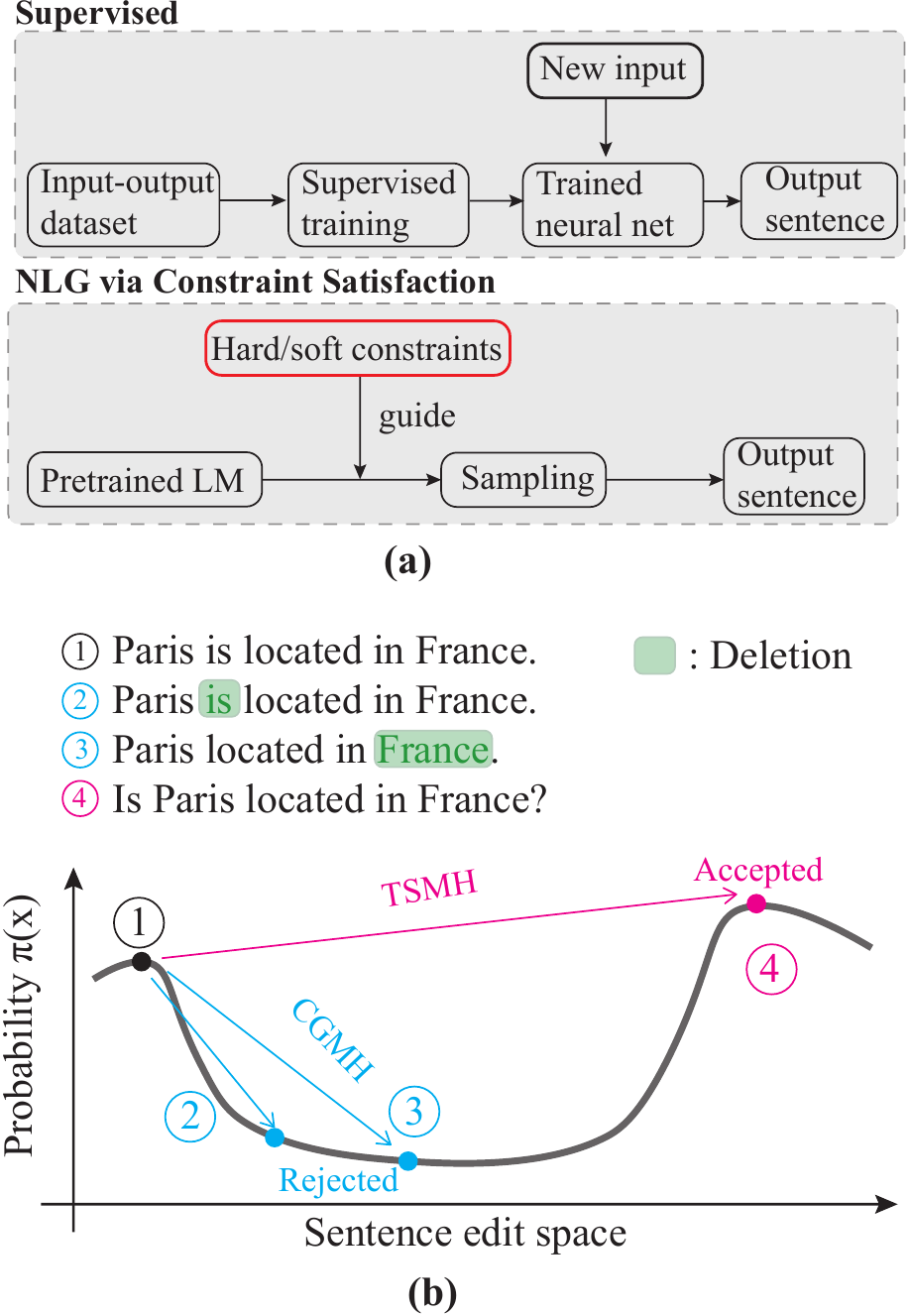}
    \vspace{-0.2cm}
    \caption{\textbf{(a)} 
    Natural language generation via constraint satisfaction (bottom), 
   comparing to supervised approach (up). \textbf{(b)} Our proposed tree search enhanced MCMC (\method, pink line) traverses the probabilistic space of high-quality sentences more effectively than the baseline (blue line). 
    }
    \label{fig:tsmc-comp}
    % \vspace{-0.5cm}
\end{figure}

Supervised techniques still dominate in natural language generation tasks. 
Despite its success, supervised approaches need to be trained with massive datasets of input-output pairs, which is non-trivial to acquire. 
In addition, it is hard to guarantee that the output sentences satisfy constraints. 
% For example, the grammar of some specific type of sentences, or the sentiment of the output sentences.
Recent approaches first pre-train a language model on a general-purpose dataset, then fine-tune the neural net on a task-specific dataset~\citep{devlin2019bert, radford2019language}. 
These approaches partially mitigate data hunger in training large and flexible neural networks. Nevertheless, they still require carefully crafted datasets for fine-tuning. 
% \zms{do they really mitigate the data hunger? they mainly improve performance...}

% We present a constraint satisfaction driven approach for language generation. In particular, we sample sentences from a probabilistic distribution proportional to the likelihood of a fixed language model measuring quality, while satisfying task-specific combinatorial constraints.
We present a constraint satisfaction driven approach for language generation. In particular, we sample sentences that attain high likelihoods from a language model and satisfy task-specific constraints.
Sampling sentences that attain high likelihoods in the language model ensures the quality of the generated sentence. 
Constraints guarantee that the sentences fit the specific language task.
The constraints can be hard ones such as the grammar rules, or soft ones such as attaining positive sentiment scores. 
%
%As an example, we consider generating imperative sentences, which must be natural (attaining high scores from a language model), 
% containing keywords (hard constraint), while having a positive sentiment score from a fixed sentiment classifier (soft constraint). 
%. In that task, we sample sentences that satisfy the grammatical rules of being imperative, containing keywords, as well as attaining high scores from the language model. 
%

Our method harnesses constraint satisfaction, rather than learning, to guide language generation. % towards specific language generation tasks. 
In fact, there is no task-specific training in our approach. %What we need is just a language model pre-trained with unlabeled general-purpose corpus and a set of pre-defined constraints. with paired input-output data points 
Our approach is highly flexible since constraints can be switched quickly to be adapted to a different task, even faster than fine-tuning. 
It also allows us to leverage the latest developments of automated reasoning for language generation. 
%
% is fundamentally different from supervised learning methods, as it requires minimal supervision. It is not required to train any model with paired input-output data supervision. What we need is just a language model pre-trained with unlabeled general-purpose corpus and a set of pre-defined constraints.
%
Although the field of language generation is dominated by learning, reasoning should play an equally important role. 
Human beings can write beautiful words from  
reasoning over what is needed in the specific writing task, without learning from previous examples. % at all. 

To better handle the combinatorial constraints, a tree search is embedded into the proposal process of the Markov chain Monte Carlo (MCMC) for constrained language generation, %\citep{andrieu2003introduction}
which suggests candidate proposals that satisfy more constraints.
Our approach is motivated by Sample-Search~\citep{DBLP:conf/aaai/GogateD07, DBLP:journals/jmlr/GogateD07, DBLP:journals/ai/GogateD11}, which integrates backtrack search into importance sampling. 
Making multiple word-level changes within one proposal step of MCMC allows the direct transition between legitimate sentences, while previous approaches must go through infeasible intermediate states. Such moves are typically rejected by MCMC and therefore result in a slow mixing rate (See Figure~\ref{fig:tsmc-comp}(b) and Section~\ref{sec-motivation}).

In literature, constrained language generation has been attacked in a supervised way in \citep{NIPS2014_5346, DBLP:conf/nips/BerglundRHKVK15, DBLP:conf/icml/HuYLSX17, DBLP:conf/acl/ZhangZML19,DBLP:conf/acl/MiaoSZL20}. 
There are also multiple works of literature which model language rules as 
decomposed tree structures ~\citep{DBLP:conf/aaai/LeeMWTC19} or sentiment tags~\citep{DBLP:conf/aaai/SuXQH18}. 
Markov Logic network~\citep{DBLP:journals/ml/RichardsonD06,DBLP:conf/emnlp/KhotBGSCE15} are also used to formulate grammar rules. 
The distance between vectors representing sentences meaning is considered as soft constraints in~\citep{DBLP:conf/acl/TsvetkovBSP18,DBLP:conf/icml/BelangerM16, amato2010sentence}. 
In a nutshell, we summarize our contributions as follows: 
\begin{enumerate}
    \item We define the problem of constraint satisfaction driven natural language generation, and propose a sampling-based approach to tackle the problem with combinatorial constraints. 

    \item We propose a Tree Search enhanced Metropolis-Hastings approach (\method) for the proposed task, which mixes faster than standard MCMC in the presence of combinatorial constraints.

    \item Experiment results on generating interrogative, imperative sentences with keywords, and sentences with given sentiments demonstrate that our \method is able to generate sentences that satisfy more hard and soft constraints as well as retain good quality.
\end{enumerate}

%% file: 2-constr.tex
\section{Language Generation via 
         Combinatorial Constraint Satisfaction}
\label{section2}
We provide a general framework for the constrained natural language generation. 
In this framework, sentences are generated by sampling from a probability distribution that is proportional to the score of a pre-trained language model times the constraint score. Formally, 
let $x$ be a sentence, $\pi(x)$ be the probability that $x$ is sampled, then $\pi(x)$ should be: 
\begin{equation}
\pi(x)\propto P_{\mathrm{LM}}(x)\cdot \mathrm{Constraint}(x).
\label{eq:pi}
\end{equation}
Here, $P_{\mathrm{LM}}(x)$ is the score of a language model \citep{DBLP:conf/interspeech/SundermeyerSN12,radford2019language}, 
which measures the quality of sentence $x$. 
Higher $P_{\mathrm{LM}}(x)$ means the sentence $x$ is better in quality. 

$\mathrm{Constraint}(x)$ is a task-specific penalty term. 
For example, in interrogative sentences generation,
we would enforce $\mathrm{Constraint}(x)$ to guarantee that
only sentences in the interrogative form receive high scores. 
Constraints are composed of hard and soft constraint terms:
\begin{equation}
    \mathrm{Constraint}(x)=\Phi_{\mathrm{hard}}(x)\cdot \Phi_{\mathrm{soft}}(x).
\end{equation}

Both the hard constraint score $\Phi_{\mathrm{hard}}(x)$ and the soft constraint score $\Phi_{\mathrm{soft}}(x)$ are float values ranging from $0$ to $1$. The closer to $1$, the more satisfied the constraints are.

Unlike supervised methods which need to be trained with paired input-output data, 
our framework can solve language generation tasks without task-specific training. 
$P_{\mathrm{LM}}(x)$ comes from a language model, only trained on general-purpose language tasks. There is no fine-tuning of  $P_{\mathrm{LM}}(x)$ on the specific task.
$\Phi_{\mathrm{hard}}(x)$ is based on crafted constraints. $\Phi_{\mathrm{soft}}(x)$ comes from either user-defined functions, or pre-trained neural networks, which again is not fine-tuned on the specific task. 
The overall formulation composed of the language model and the task-specific constraints allows us to sample sentences 
which are close to natural language while satisfying constraints.

\subsection{Hard Constraints}

In this paper, we use propositional logic to define hard constraints $\Phi_{\mathrm{hard}}(x)$.
Nevertheless, our sampling approach generalizes to other logic forms. 
We leave the generalization to first-order logic as future work. 
For hard constraints, we define $\Phi_{\mathrm{hard}}(x)$ as 
\begin{equation}
\Phi_{\mathrm{hard}}(x)=\beta^{M-\sum_i c_i(x)}
\end{equation}
where $c_i(x)$ is an indicator variable which takes 1 if the sentence $x$
satisfies the $i$-th constraint, and $M$ is the total number of hard constraints. 
$\beta$ is between 0 and 1. 
We use quite small $\beta$ values in our experiments, which put a large penalty on violating one hard constraint. 
We also define \textit{Constraint Error} $C(x)$ as the number of hard constraints a sentence violates, i.e., $C(x) = M - \sum_i c_i(x)$.
Constraints are defined in the logical form of word categories. 

\noindent\textbf{Word Category Division}~~We divide the entire vocabulary into several categories of words.\label{word-category-def} Given vocabulary set $U$, we partition $U$ into non-overlapping subsets:
$\mathcal{V}=\{V_1, V_2, \dots, V_{|\mathcal{V}|}\}$, satisfying: (i) all $V_i$ are subsets of $U$: $V_i \subseteq U, \ \ \forall V_i \in \mathcal{V}$; (ii) categories are non-overlapping: $V_i \cap V_j = \varnothing, \ \ \forall V_i,V_j \in \mathcal{V},i\neq j$; (iii) $V_i$ together cover the whole vocabulary: $\bigcup_i^{|\mathcal{V}|} V_i = U$.

The word category division strategy varies for different tasks.
For example, we split the whole vocabulary into $\mathcal{V}=\{\texttt{[QWH]}, \texttt{[AUX]}, \texttt{[OTH]}\}$
for generating interrogative sentences. 
Here, $V_1=$\texttt{[QWH]} represents the set of \textit{wh}-words leading a question: \textit{what, when, where, which, who, whom, whose, why, how}.  $V_2=$\texttt{[AUX]} represents the set of auxiliary verbs and copula words: \textit{do, does, did, be, am, are, is, \dots,} etc. $V_3=\texttt{[OTH]}$ means all other words in the vocabulary. 
% \textit{do, does, did, be, am, are, is, was, were, shall, will, should, would, can, could, may, might, must}
%
We may use another division in, e.g., generating imperative sentences. 
Sometimes we need to generate sentences with keywords. We let each keyword forms a category.
For example, to generate interrogative sentences with the keyword \textit{learning},  the division would be:
$\mathcal{V}=\{\texttt{[QWH]}, \texttt{[AUX]}, [\textit{learning}], \texttt{[OTH]}\}$.

\noindent\textbf{Hard Constraints}~~Given a sentence with length $m$ \footnote{As we conduct sampling for the sentence, sentence length is pre-known and we set $m$ as the length of the longest one.
}, 
let $w_i^{V_j} \in \{true, false\}$ be an indicator variable
that the  $i$-th  word in the sentence is in category $V_j$. For example, variable $w_1^{\texttt{[QWH]}}=true$  if and only if the first word in sentence is a \textit{wh}-like word.
For sentence-level constraints, we can define them using
propositional logic over $w_i^{V_j}$ (\texttt{and} ($\wedge$), \texttt{or} ($\vee$), \texttt{not} ($\neg$)). 
We give a few examples below.

\noindent\textit{Enforcing Keywords in a Sentence}~~
Given one keyword $\texttt{K}$, we can enforce its existence in the sentence using the following constraint:
\begin{equation*}%\label{lab:key}
w_1^{\texttt{[K]}} \vee w_2^{\texttt{[K]}} \vee \dots \vee w_m^{\texttt{[K]}}.
\end{equation*}
here $\texttt{[K]}$ is a set containing the  keyword $\texttt{K}$. 
We formulate this constraint assuming a known sentence length $m$. Indeed, length $m$ is a variable and can vary over the sampling procedure. Nevertheless, as we can see shortly in the sampling process,  the lengths are known for both sentences when transiting from one sentence to another. Therefore, the semantic meaning of $m$ is clear during  sampling. Details on the sampling process is in Section~\ref{sec-algo}.

\noindent\textit{Enforcing Imperative Sentence}~~
According to the definition in \citep{aarts1989imperative}, the starting word of an imperative sentence should be either a verb: $w_1^{\texttt{[VERB]}}$ or an adverb followed by a verb:  $w_1^{\texttt{[ADV]}} \wedge w_2^{\texttt{[VERB]}}$. We encode such constraint as: 
\begin{equation*}%\label{lab:imper}
w_1^{\texttt{[VERB]}} \vee (w_1^{\texttt{[ADV]}} \wedge w_2^{\texttt{[VERB]}}).
\end{equation*}

\noindent\textit{Enforcing Interrogative Sentence}~~
\label{qg-constraints}
We use the following two constraints to enforce the sentence to be 
interrogative:
(i) The first word is in \texttt{[QWH]}. (ii) The second or third word in the sentence is in \texttt{[AUX]}.
(i, ii) can be written together as:
\begin{equation*}%\label{lab:intro}
{\scriptstyle w_1^{\texttt{[QWH]}}\wedge \left((w_2^{\texttt{[AUX]}}\wedge \neg w_3^{\texttt{[AUX]}}) \vee ( w_3^{\texttt{[AUX]}}\wedge \neg w_2^{\texttt{[AUX]}})\right)}.
\end{equation*}

This constraint is similar to the definition in \citep{DBLP:conf/ijcai/ZhangQYYZ17}. 
We acknowledge that this is a relaxed constraint.
% in terms of the interrogative sentence grammar. 
%
Nevertheless, our sampling approach also consider the score from language model. These constraints accompanied
with the language model guide us to good interrogative sentences in practice. 

\subsection{Soft Constraints}
\label{sec-soft-constraint}
A soft constraint  assigns a float value between $0$ and $1$ to indicate how the constraint is satisfied.  
For tasks with only hard constraints, $\Phi_{\mathrm{soft}}(x)$ is set to $1.0$.
Soft constraints can be derived quite flexibly. It can be from a user-defined function (see ``sentence similarity'' for an example), or from a pre-trained neural network (see ``sentiment score''):

\noindent\textit{Sentence Similarity}
We can define a soft constraint function ensuring that the generated sentence $x$ is close to the reference sentence $y$ in semantic meaning. 
For one word in sentence $x$, we first find the closest word in sentence $y$ by computing their cosine similarity. Then either the minimum or the average of these words' cosine similarity is taken as the similarity score for sentence $x$ and $y$.

\noindent\textit{Sentiment Score} We can enforce that the generated sentence must have a given sentiment by enforcing the value for the sentence from a sentiment analysis model.
The output score of a sentiment analysis neural net represents whether the sentence has a positive or negative sentiment. We use this score as a soft constraint to control the sentiment of generated sentence with positive or negative attitude. Notice that the sentiment analysis neural net is pre-trained
on a separate dataset and remains intact in our framework.

This setup gives us additional flexibility. To be specific, if we need to generate sentences that contain keywords while having a given sentiment, it is difficult to find a large dataset of this type and the performance of a pure learning approach may be limited. 
To summarize, the main attribute of the constraint satisfaction framework is allowing a formulation using both hard and soft constraints, without the need of task-specific training or tuning. 

%% file: 3-sample.tex
\section{Tree Search Enhanced MCMC}

\begin{figure*}[!htb]
\centering
\includegraphics[width=0.95\linewidth]{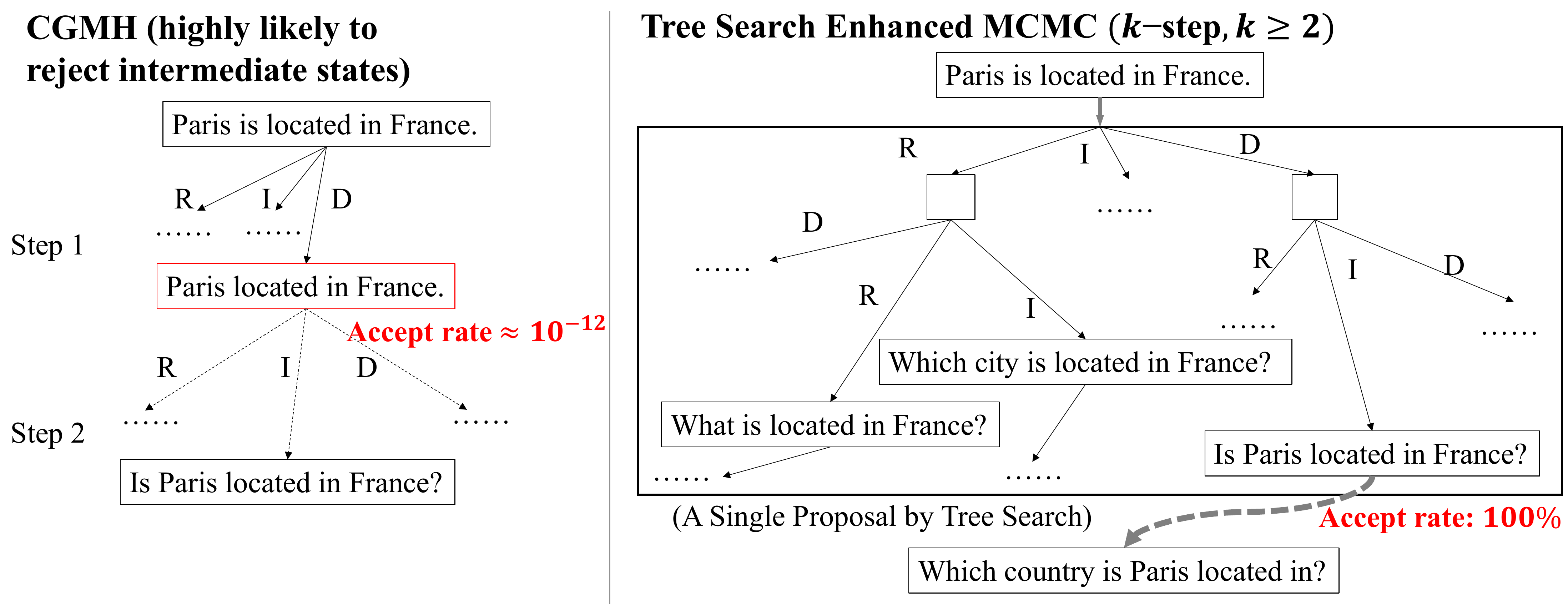}
\caption{Our method, tree search embedded MCMC (\method), outperforms 
CGMH in generating sentences with complex combinatorial constraints. 
(Left) CGMH must pass intermediate sentence states (highlighted in red), which have very low acceptance rate to reach the intermediate sentence \textit{Is Paris located in France?} starting from sentence \textit{Paris is located in France}. 
This results in the poor performance of CGMH when handling combinatorial constraints. 
(Right) By embedding a tree search into MCMC, \method can reach the 
an intermediate sentence from the starting sentence in one step, and with 
an acceptance rate of 100\%. 
R, I, D mean \textit{replace, insert, delete}. See Section~\ref{sec-motivation} for a detailed discussion.}
 \label{fig:search-proc-1}
\end{figure*}

Markov chain Monte Carlo (MCMC) is a classical approach to 
sample sentences from probability distribution $\pi(x)$ as defined in Equation~\ref{eq:pi}. 
Starting from one sentence $x$, MCMC moves to the next sentence $x^*$
by first generating a sample $x^*$ from the proposal distribution $Q(x^*|x)$ and then accept $x^*$ with the following acceptance rate $A(x^*|x)$:
\begin{equation}\label{eq:acc}
A(x^*|x)
=\min\left\{1, \frac{\pi(x^*)Q(x|x^*)}{\pi(x)Q(x^*|x)}\right\},
\end{equation}
If sentence $x^*$ is rejected, then the sample remains at $x$. The distribution of samples will converge to the sentence stationary distribution of Markov chain $\pi(x)$.
Previous work~\citep{miao2019cgmh} proposes to use MCMC for constrained sentence generation, namely CGMH algorithm. Their proposal distribution only suggests sentences with one-word modification.
Nevertheless, CGMH cannot handle the combinatorial constraints in our problem definition, because of the \textit{low acceptance ratio problem} caused by the \textit{locality} of the proposal distribution. 
In other words, the sampling process can only visit a limited number of neighbors, thus the Markov chain will easily be trapped at one infeasible state, resulting in a lot of  rejections.
We illustrate this problem in detail and hence motivate our tree search embedded MCMC approach using the following example.

\subsection{Motivation: Breaking the low  acceptance barrier}

\label{sec-motivation}

Suppose we need to
generate a question, whose answer comes from an underlined part of a sentence. For example, suppose we underline \textit{France} in the sentence: 

\textit{A: Paris is located in \underline{France}.}

\noindent The question we would like to generate is: 

\textit{B: Which country is Paris located in?}

Under our constraint satisfaction framework, we define $\mathrm{Constraint}(x)$ so that 
real interrogative sentences 
such as question \textit{B} would receive high probability in the defined  $\pi(x)$. 
Our constraints are: (i) the whole sentence is in the interrogative form. 
% (ii) \textit{France} cannot appear in the sentence. 
(ii) \textit{Paris} and \textit{located} must appear
in the sentence. 
We run MCMC starting from sentence \textit{A}. % then output the sentence that gets the high probability in the MCMC chain.

It is hard for MCMC without tree search to generate question \textit{B} in a reasonable time starting from \textit{A}. 
Because the edit distance between sentence \textit{A} and \textit{B} is larger than $2$, we cannot generate \textit{B} from \textit{A} with one step of word insertion, removal, or replacement. 
In order for CGMH to reach \textit{B} from \textit{A}, it has to 
encounter a few intermediate steps. Without loss of generality, suppose CGMH proposes sentence \textit{C} in one MCMC step 
by removing \textit{is}:

\textit{C: Paris \st{is} located in France.}

Notice that \textit{C} is not a legitimate English sentence, so its language model score $P_{\mathrm{LM}}(x)$ becomes much
smaller compared to the original sentence \textit{A}. 
In addition, \textit{C} violates more constraints than \textit{A}, which decreases its $\mathrm{Constraint}(x)$ score as well. 
In MCMC, the probability to accept the move from \textit{A}  to
sentence \textit{C} is given by 
Equation~\ref{eq:acc}, in which the dominating term is 
$\frac{\pi(C)}{\pi(A)} = \frac{P_{\mathrm{LM}}(C)~ \mathrm{Constraint}(C)}{P_{\mathrm{LM}}(A)~\mathrm{Constraint}(A)}$. 
Because both $P_{\mathrm{LM}}(C)$ and $\mathrm{Constraint}(C)$ are smaller, the acceptance ratio becomes really small. In fact,
we found the acceptance ratio to be $5.93\times 10^{-12}$ in
our experiment. 
This means that it will take CGMH many steps (on the order of
$10^{12}$) to move one step from sentence \textit{A} to \textit{C}. 
Figure~\ref{fig:search-proc-1} (left) demonstrates this. It is easy to verify that barriers of low acceptance rate exist on every path from sentence \textit{A} to \textit{C} and thus the rejection problem exists. 

On the other hand, 
if we allow the proposal distribution to suggest sentences with multiple  word-level changes, one can transit from sentence \textit{A} to \textit{B} through all legitimate sentences as intermediate steps. Consider the following two-step change:
\begin{enumerate}
    \item First delete \textit{is} and insert \textit{is} before \textit{Paris}. This changes sentence \textit{A} to \textit{D:\\ {Is Paris located in France?}}
    \item Delete \textit{France} and insert \textit{Which} and \textit{country}. This changes sentence \textit{D} to \textit{B}.
\end{enumerate}

Because the intermediate step sentence \textit{D} is a legitimate English sentence and $\mathrm{Constraint}(D)=\mathrm{Constraint}(A)$,  $\frac{\pi(D)}{\pi(A)}$ is close to $1$, resulting in %and therefore leading to
a $100\%$ acceptance ratio in this step. When changing from $D$ to $B$, 
notice that $B$ is also a legitimate sentence and 
it satisfies more constraints than $D$. In fact, the acceptance ratio is also $100\%$. %in this step. 
Figure~\ref{fig:search-proc-1} (right) demonstrates this case. 

For tasks with soft constraints, there are also similar rejection problems for CGMH. For example, \textit{``Nothing is impossible"} is a sentence with positive sentiment. If we insert, replace or delete one word, it is hard to keep the sentence valid and preserve the positive sentiment. %  

Motivated by these examples, we propose the embed a tree search into the proposal process of MCMC to solve the rejection problem, which suggests candidate sentences with multiple word-level changes and satisfy more constraints. 

%% file: 4-search.tex
\subsection{TSMH Algorithm Implementation}
\label{sec-algo}
Our Tree Search enhanced Metropolis-Hastings (TSMH) still follows the classical MCMC procedure. The only difference is a 
new proposal distribution $Q(x^*|x)$ generated from a tree search process. 
The tree search defines a probability distribution over \textit{templates} of sentence moves. 
Each template defines a subset of possible moves. 
The sentences within the same template satisfy the same hard constraints score $\Phi_{\mathrm{hard}}(x)$. 
The proposal probability distribution induced by the tree search algorithm biases towards templates that have high $\mathrm{Constraint}(x)$ scores. 

\label{template-def} 
A \textbf{template} defines a set of sentences where each word is either given or specified by a word category. 
For example, a template \texttt{[[QWH]},\texttt{[AUX]},\texttt{[OTH]},\texttt{[OTH]]} restricts that the first word must be a \textit{wh}-word, the second word must be an auxiliary verb and the last two words must be other words. 

Notice that we can decide how many hard constraints a sentence satisfies at the template level, since the indicator variables in the constraints defined in this paper only restrict the categories of words. For example, the template \texttt{[[QWH],[AUX],[OTH],[OTH]]} satisfies the constraints of being an interrogative sentence defined in Section \ref{section2}. 
Our proposal procedure first sample a template and then fills in this template
with words based on a language model.

\noindent\textbf{Overview of the Proposal Process}~~ During the sampling process,
suppose we are at one sentence $x$. We will
sample a new sentence $x^*$ from the proposal distribution as follows. 
First, our algorithm will decide the \textit{\textbf{positions}} of the words to change by random selection. Typically, our algorithm will change more than one word. 
Then we use a tree \textit{\textbf{search}}, which enumerates all possible operations on 
the selected words. This includes deciding the operation on each word (\textit{insert, delete, or replace}) as well as the associated word category in case of \textit{insert} and \textit{replacement}. 
In this case, every leaf branch of the search tree will be a sentence template. 
Because the number of word categories is limited, the tree search procedure is often cheap.
As discussed, we can infer the number of hard constraints satisfied based on the template associated with each tree leaf. 
We then \textit{\textbf{rank}} these templates based on the number of constraints satisfied and sample one template based on a geometric series, favoring templates
that satisfy more constraints. 
Finally, we \textit{\textbf{fill}} in the sampled template with words suggested by a language model, and then select one filled sentence $\hat{x}$ as proposal, according to the language model score times the soft constraint score $P_{\mathrm{LM}}(\hat{x})\cdot \Phi_{\mathrm{soft}}(\hat{x})$.
% \zms{talk about soft} 
% For soft constraints, it is considered as the $\mathrm{Constraint_{soft}}(x)$ score is included when computing the stationary probability $\pi(x)$ and the acceptance rate $A(x^*|x)$. \zms{it is used in the step of template selection}
%
Soft constraints $\Phi_{\mathrm{soft}}(x)$ give us a real number, which is similar to the language model $P_{\mathrm{LM}}(x)$. 
We treat them together with the language model in the proposal process. 

Our approach alleviates the rejection problem of MCMC by enumerating all possibilities in the space of multiple word change at the template level, based on the analysis in section \ref{sec-motivation}. 
%
% Suppose there is a word changing plan that satisfies many constraints, our tree search \textit{will} be able to find it and our proposal distribution will favor such a plan. 
%
This process enables us to handle combinatorial constraints. 
Tree search also allows us to prune useless branches.
% During the search process, some branches may already violate more constraints than the current best template.
%

% Since we use a geometric distribution that decays very fast with the increasing number of constraint violations, 
% we will prune such branches early because of the small probability that any template under the branch is selected. 

\begin{figure}[t]
\centering
\includegraphics[width=1.0\columnwidth]{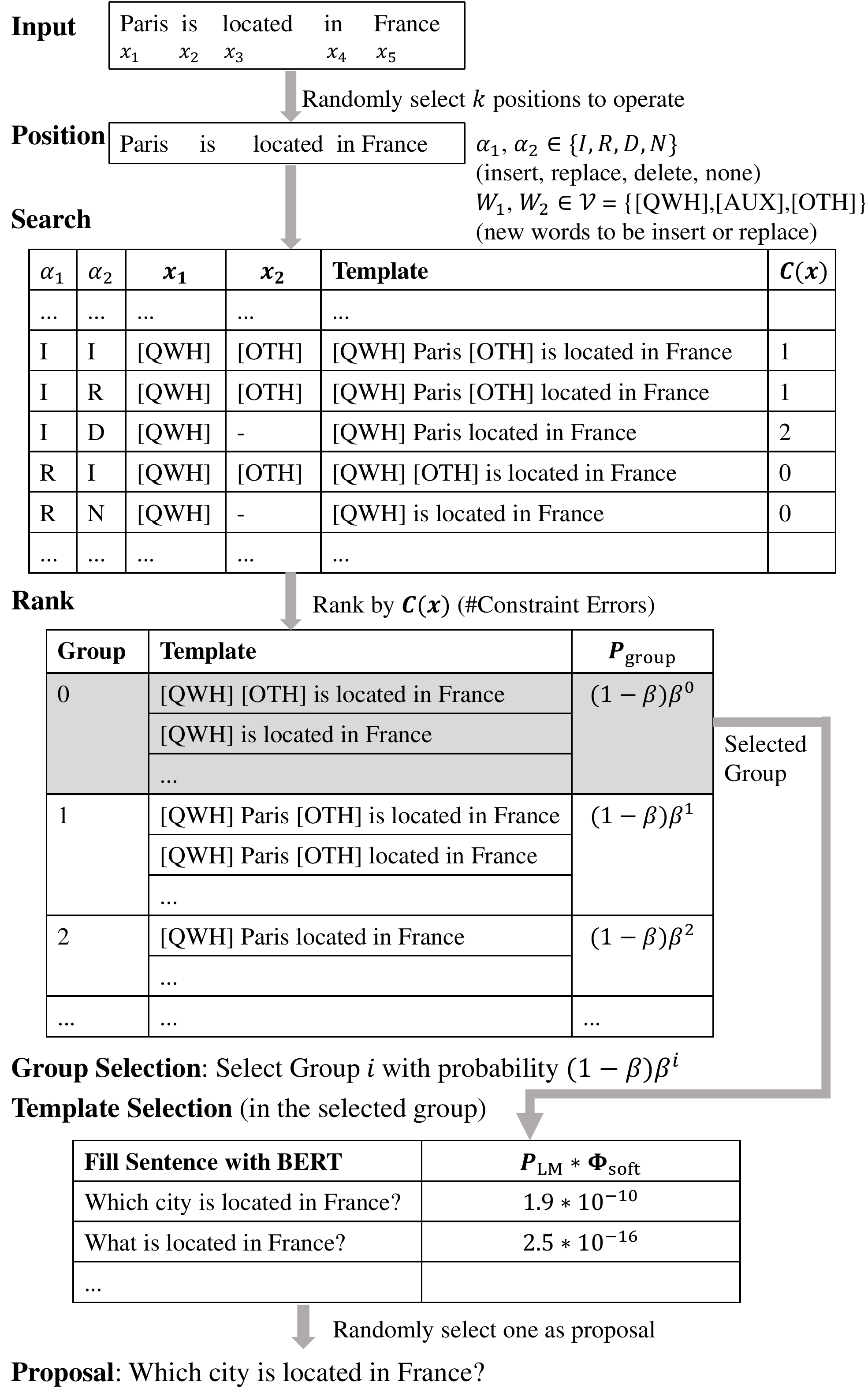}
\caption{The proposal process of Tree Search Embedded MCMC. The input is the current sentence (state) and the output is the proposed sentence. This proposal process favors
sentences satisfying a large number of constraints.}
\label{figure-search-proc}
\end{figure}

\subsubsection{Detailed Search Procedure}
The procedure of searching proposals in our tree search embedded MCMC is as follows and shown in Figure~\ref{figure-search-proc}. \label{detail-search-proc}

\noindent \textbf{Position} Randomly select $k$ positions $\{t_1,\dots, t_k\}$ to perform word-level operations with uniform probabilities, where $k$ is the size of the search steps.
    The probability of getting each combination of positions is: $P_{\mathrm{pos}}=1/{\binom{m}{k}} $,
    where $m$ is the length of the sentence.

\noindent\textbf{Search} Search and iterate all different operations and all different word categories (mentioned in Section \ref{word-category-def}) for each selected position. For example, if we have $|\mathcal{V}|$ word categories and the operation set \textit{\{replace, insert, delete, none\}}
, we need to enumerate $(2|\mathcal{V}|+2)^k$ different combinations of operations and word categories. We use word placeholders \texttt{[MASK]} to represent the unknown inserted or replaced words. We keep track of all the generated \textbf{templates}  and their corresponding numbers of violated constraints. 

\noindent\textbf{Rank and Group Selection} 
We define a group as the set of templates which violate the same number of constraints.
We sort all templates by its number of violated constraints (constraint error) $C$ in ascending order, and put templates with the same $C$ into one group. 
We then randomly select group $i$ with probability:
%\begin{equation}
$P_{\mathrm{group}}=(1-\beta) \cdot \beta^{C_i-\min_j{C_j}}$, 
%\end{equation}
where $C_i$ is the constraint error of group $i$, and 
$\beta$ is a very small float value (like $10^{-10}$).
In this way, we favor choosing the group satisfying the largest amount of constraints, while also ensuring the irreducibility of the Markov chain. Let the chosen group at this step be $G_i$.

\noindent\textbf{Fill and Template Selection} 
In this step we will first fill every template with words in the selected group $G_i$, then we select one filled template as the proposal. 
Because the template restricts the masked word to be chosen only from the corresponding word category, we \textit{\textbf{fill}} it by selecting words from the given word category.
The probability of selecting the $t_i$-th word $P_{{\mathrm{Fill}}_i}$ is the conditional probability of filling words at this locations given contexts:
$P_{\mathrm{LM}}(x_{t_i}|x_1, ..., x_{t_i-1}, x_{t_i+1}, ..., x_m)$.
% if we need to \textit{replace} or \textit{insert} a word at location $t_i$.
% For each template in the selected group, we first \textbf{fill} it by sampling the unknown masked words from the subset of vocabulary indicated by the template, which are inserted or replaced as proposal and represented by word categories in the templates, according to the conditional probability of filling words at those locations.
% \xyx{the previous sentence is very long and confusing} 
The probability of getting one sampled sentence is: 
$P_{\mathrm{fill}}=\prod_{i=1}^k {P_{{\mathrm{fill}}_i}}$, where $i$ means the word level action for $i$-th position we selected. If the operation in $t_i$ is \textit{delete} or \textit{none}, then $P_{\mathrm{fill}_i}=1$. 
% The term $P_{\mathrm{LM}}(x_{t_i}|x_1, ..., x_{t_i-1}, x_{t_i+1}, ..., x_m)$ can be evaluated with bidirectional language model by replacing $x_{t_i}$ with a mask token~\citep{devlin2019bert}, or approximated with a normal language model by pre-selecting $x_{t_i}$ candidates and normalize $P_{\mathrm{LM}}(x_{t_i}|x_1, ..., x_{t_i}, ..., x_m)$ over the selected candidates, which has been used in ~\citep{miao2019cgmh}.
%
% For each template in the selected group, we compute the sampled sentence' probability $P_{\mathrm{LM}}(x_1,\dots, x_m)$ by the language models. Then 
We sample one template within the group (together with the corresponding sampled sentence) according to the sentence probability times soft constraint score: 
$P_{\mathrm{template}}=\frac{P_{\mathrm{LM}}(x^*)\cdot \Phi_{\mathrm{soft}}(x^*)}{\sum_{\hat{x}\in G_i} P_{\mathrm{LM}}(\hat{x})\cdot \Phi_{\mathrm{soft}}(\hat{x})}
$.  % use equation to express normalize
% where norm means normalize the probabilities within the selected group and $G_i$ means all templates in the group.
%

The proposal distribution $Q(x^*|x)$  leading from sentence state $x$ to $x^*$ in this procedure is
% \xyx{confusing! What is $G_i$?? First selection or first fill?? The group is chosen in the previous step, isn't it? What is template $P_{temp}$ then?}
%\begin{align}
%\begin{split}
$Q(x^*|x)=P_{\mathrm{pos}}P_{\mathrm{group}}P_{\mathrm{fill}}P_{\mathrm{template}}$.
%\end{split}
%\end{align}

%% file: 5-exp.tex
\input{5.1.table1.tex}

\section{Experiments}
We evaluate our approach on three applications: interrogative, imperative, and fixed sentiment sentences generation. 
In each task, we construct the specified type of sentences by sampling starting from keywords and enforcing task-specific constraints. 
For each task, we run our \method algorithm for 100 steps, with 100 candidate sentences generated. $k$ is set to 3. %tree search MCMC algorithm to generate 100 samples.
Since the tree search in \method considers changing 3 words at each iteration, we run the baseline CGMH for 300 steps as a comparison.
We select the sentence with the highest $\pi(x)$ value among the sentences generated by each algorithm as the output.
Our results are summarized in Table \ref{table-results}.

In general, our method \method outperforms baselines and generates sentences that satisfy more constraints, are of good quality and are likely to be close to the natural language.
Our main results are summarized in Table \ref{table-results}, in which Valid\% denotes the percentage of generated sentences 
that satisfy all constraints. $\pi(x)$ is the value of the stationary probability $P_{\mathrm{LM}}(x) \cdot \mathrm{Constraint}(x)$. 
$P_{\mathrm{GPT-2}}(x)$ is language model probability estimated by a pre-trained GPT-2 model, which measures the quality of the sentences. 
Accept\% means the acceptance rate of MCMC. 
Detailed experiment settings can be reviewed in appendix~\ref{sup-exp-settting}.

\subsection{Interrogative Sentence Generation}
\label{exp-ques}
In the interrogative sentence generation, we construct interrogative sentences by sampling starting from the keywords. 
We enforce that sentences with a high probability to be sampled must satisfy grammar constraints of being interrogative and contain a few given keywords. 
The constraint definition for interrogative sentences is in section \ref{qg-constraints}.

According to the results, in the experiment with keywords, 92.67\% of the output sentences of our \method algorithm satisfy all the constraints, while merely 18.33\% satisfy constraints for the baseline. The numbers are 83.17\% 
and 45.50\% for the experiment without keywords, respectively. 
This demonstrates that our \method generates sentences with more constraints satisfied.
In addition, our method has a higher $\pi(x)$ (stationary probability value) and acceptance rate, suggesting that the tree search embedded help MCMC to mix faster.
Overall, our method \method can handle more complicated constraints in language generation tasks. 
%
% We also compare with another method~\citep{lewis2019unsupervisedqa} and we put the results in the appendix~\ref{sup-compare-uqa}. Notice the comparison may not be entirely fair, because the experimental settings are different.
%
\begin{table}[!ht]
    \centering
    \begin{tabular}{l|cc}\toprule
   Methods & \#Votes &  Votes\%\\\midrule
        CGMH & 196 & 33.64\% 
       \\
       \method (Ours) &\textbf{384}  & \textbf{66.36}\%  \\
         \bottomrule
    \end{tabular}
    \caption{Human evaluation of the quality of the generated interrogative sentences from keywords in terms of fluency and grammar. Most human participants (native speakers) agree that the sentences 
    generated by our \method are better in quality compared to CGMH.}
    \label{tab:human}
\end{table}

% \noindent\textbf{Human Evaluation}~~We conduct a human evaluation to see which approach generates sentences with better quality. We evaluate on  interrogative sentences generation with keywords. 
%
% As shown in Table~\ref{tab:human}, sentences from our model receive almost twice many votes of the baseline, which suggests that the sentences generated by our approach are better in quality. 
% %
% Details of our human evaluation can be found in the appendix~\ref{sup-human-eval}.

\noindent\textbf{Human Evaluation}~~We conduct human evaluation for  interrogative sentences generated with keywords. 
We present human participants from the Amazon Mechanical Turk with a pair
of sentences at a time. One sentence is generated by our \method model and the other one is from the baseline CGMH. 
We ask human participants which sentence is better in terms of fluency and grammar. 
%
% In this evaluation, $32.72\%$ of the participants agree that our \method model generates better sentences. 
%
In terms of the experimental setting, we use 100 sentence pairs generated by CGMH and \method with the same keyword inputs. We randomly split these 100 test sentence pairs into 5 survey groups, and then deploy them on the Amazon Mechanical Turk.
We randomly assign human participants to survey groups.
%for native speakers to vote which one is better. 
When showing the sentence pairs, we also provide the keywords that the sentences 
must contain. 
We ask human participants to vote which sentence in the pair is better in terms of grammar coherence, keyword coverage and fluency. 
We use a gold-standard question to detect if the voter is randomly doing the survey. 
Every valid survey contains a randomized set of 20 questions. We received in all 580 votes. Each question pair receives votes ranging from $5$ to $11$. As shown in Table~\ref{tab:human}, sentences from our model receive almost twice times of votes than the baseline, which suggests that the sentences generated by our approach is better in human evaluation.

\input{5.2.table2}
\noindent \textbf{Case Studies} As shown in Table~\ref{tab:case}, we compare some output sentences of our method with the baseline using the same inputs and keywords. More examples can be seen in the appendix~\ref{sup-case-study}.
From these cases, we can see that our method generates sentences with better quality.

\noindent \textbf{Comparison with Other Methods} 
\label{sup-compare-uqa}
We compare our \method method with UQA~\citep{lewis2019unsupervisedqa}. The setting of UQA is different from us: it takes a paragraph as input and generates a corresponding question.
% UQA also cannot enforce keyword constraints, and it is trained with a corpus of interrogative sentences. 
Although this comparison is not fair, the baseline is the most similar and the best framework that we can compare with. To run UQA, we use the corresponding original sentences from which the keywords of \method are extracted as the input. In other words, for \method, the inputs are keywords extracted from the SQuAD 2.0~\cite{DBLP:conf/acl/RajpurkarJL18} questions. For UQA, we take the corresponding paragraphs of the selected questions as input. This also gives UQA additional advantage because it has access to a paragraph, rather than keywords. To make it more comparable, we remove the keyword constraints in this experiment.  In Table~\ref{tab:exp-uqa}, we compare the language model scores $\log P_{\mathrm{LM}}$ of the generated sentences that reflect the naturalness and fluency, and the stationary probability $\pi(x)$ and valid percentage Valid\% that show how good it satisfies our pre-defined constraints. 
%We also detect whether the grammar of the output sentences are interrogative by applying a constituency parser from Stanford CoreNLP~\citep{manning-EtAl:2014:P14-5}.
% From the results shown in Table~\ref{tab:exp-uqa}, we can see that our \method generates relatively good sentences (high language model scores) even without any task-specific training. The sentences generated also satisfy our predefined interrogative sentence constraints better. 
%
We pointed out that UQA was trained on the specific interrogative sentences while our method was not trained at all.
%In terms of grammar metric, the UQA method performs relatively better as it 

%\begin{table}[!htb]
%\small
%\begin{tabular}{l|rrrr}
%\toprule
%Methods & $\pi(x)$ & Valid\% & $\log P_{\mathrm{LM}}$ & Grammar\% \\ \midrule
%\method & \textbf{0.0063} &  \textbf{83.17\%} &  \textbf{-58.27}     & 80.33\% \\ 
%UQA  & 0.0024   & 50\%    & -92.75     &  \textbf{95.20\%} \\ 
%\bottomrule
%\end{tabular}
%\caption{Comparison with UQA. ``Grammar\%" means the percentage of generated sentences that are detected to be interrogative by a constituency parser.}
%\label{tab:exp-uqa}
%\end{table}
\begin{table}[!htb]
%\small
\centering
\begin{tabular}{l|rrr}
\toprule
Methods & $\pi(x)$ & Valid\% & $\log P_{\mathrm{LM}}$ \\ \midrule
UQA  & 0.0024   & 50\%    & -92.75     \\ 
\method & \textbf{0.0063} &  \textbf{83.17\%} &  \textbf{-58.27}     \\ 
\bottomrule
\end{tabular}
\caption{Comparison with UQA. Our \method outperforms UQA in terms of the percentage of satisfying the interrogative sentence constraints, and has a higher score predicted by a language model, despite UQA is trained on specific interrogative sentences while our method is not trained at all.}
\label{tab:exp-uqa}
\end{table}

\subsection{Imperative Sentence Generation}
\label{exp-imp}
We generate imperative sentences via sampling starting from the keywords. 
% Our Markov chain starts from several keywords. 
We enforce grammar constraints of being an  imperative sentence: the starting word should be either a verb $w_1^{\texttt{[VERB]}}$  or an adverb followed by a verb  $w_1^{\texttt{[ADV]}} \wedge w_2^{\texttt{[VERB]}}$. % (Equation~\ref{lab:imper}). 
We also enforce keyword constraints in this task.

As shown in Table~\ref{table-results}, our method has a higher valid percentage of 97.75\% compared to 91.32\% of the baseline, showing that the sentences generated by our method can satisfy more constraints.  Our method has a higher $\pi(x)$ (stationary probability value) and acceptance rate, suggesting our approach has a better mixing behavior. 
Overall, results show that our method using Tree Search Embedded MCMC can handle more complicated combinatorial constraints in language generation.

\subsection{Sentence Generation with Given Sentiments}
In this task, we require the sentences to contain the specified keywords and have positive sentiments~\citep{fu2019rethinking}. 
We enforce the sentences to attain high scores from a sentiment analysis neural network.
We also enforce keyword constraints as hard constraints.
We need to emphasize that, our method uses a model pre-trained on a separate dataset for sentiment analysis, which is kept intact in our experiment. No additional fine-tuning to the sentiment analysis model was performed. 
%to estimate sentiment score here, the supervision required in our method is significantly less than direct supervised learning. For supervised learning, the training data needs to be the paired form like ``\textit{keywords--positive sentence containing the keywords}"; while training a sentiment analyzer only requires a corpus with positive or negative labels, which is easier to acquire.
we consider two sub-tasks in Table~\ref{table-sentiment}: (i) positive sentiment to positive sentiment (P2P), where the input keywords are extracted from sentences which originally have positive sentiments; (ii) negative sentiment to positive sentiment (N2P), where the keywords are extracted from sentences with negative sentiments.
N2P is more difficult as it requires transforming the sentiment. 

Our method has a higher sentiment score, suggesting that our method generates sentences with more positive sentiments (better aligned with the target of this experiment). 
The increase against CGMH is bigger on the more difficult N2P task, 
which requires flipping the sentiment. 
Our model also leads in terms of language model scores, suggesting the language quality is better. 

\input{5.3.table3}

\input{5.4.table4}

% We also compare our \method method with an existing constrained text generation method CtrlGen~\citep{DBLP:conf/icml/HuYLSX17} on this task and results are shown in Table~\ref{tab:ctrl-gen}. 
%
% We need to mention that CtrlGen was trained on paired input-output data \xyx{correct?} while ours were not. 
% ctrl-gen is trained like "adversarial training"
% More details of the experiment settings can be seen in appendix~\ref{sup-ctrl-gen}.
% From the results we can see that the sentences generated by our method receive higher language model scores and more positive sentiment scores, suggesting that our method outperforms CtrlGen in the quality of the sentences generated and in matching the sentiment requirement. 

\noindent \textbf{Comparison with Other Methods} 
\label{sup-ctrl-gen}
We compare our method with CtrlGen~\citep{DBLP:conf/icml/HuYLSX17}. The setting is a little different from ours: it takes a sentence with a negative sentiment as input and transforms it to positive, without the guarantee of satisfying keyword constraints.  Our method takes a set of keywords as input. To make the outputs comparable, we select the same set of negative sentences as the input of CtrlGen and extract the keywords of those sentences as the input of \method. 
% \xyx{This gives additional advantage to CtrlGen because it has access to the whole sentences of negative sentiment. Correct?}  
Our method requires no additional training besides a pre-trained sentiment analysis model and a pre-trained language model, while CtrlGen requires training the auto-encoder. 

The results in Table~\ref{tab:ctrl-gen} show that our method outperforms CtrlGen in terms of both sentence quality and sentiment, as the sentences generated by our method receive higher language model scores and sentiment scores. 
% In terms of the acceptance rate metric, the CtrlGen method tends to generate sentence; 
% In terms of the language model score, our method gains ; In terms of the sentiment score, our method can

%% file: 5.1.table1.tex
\begin{table*}[!htb]
\centering
\begin{tabular}{l|rrrrrrrr}
\toprule
Tasks&Methods& \#sample& step & Valid\%&$\pi(x)$& $P_{\mathrm{GPT-2}}(x)$ &Accept\%  \\
\midrule
\multirow{ 2}{*}{Interrogative}&CGMH&300	&1&	18.33\%&	2.60E-04&	1.78E-18&	5.45\%\\
&\method(Ours)&100	&3&	\textbf{92.67\%}&	\textbf{1.44E-03}&	\textbf{5.51E-18}&	\textbf{24.50\%}\\
 \midrule
\multirow{ 2}{*}{Imperative}&CGMH&300&	1&	91.32\%&	0.0004&	9.86E-16&	5.49\%\\
&\method(Ours)&100	&3&	\textbf{97.75\%}&	\textbf{0.0060}&	\textbf{6.60E-15}&	\textbf{15.66\%}\\
\midrule
\multirow{ 2}{*}{Sentiment}&CGMH&300&	1&	96.33\%&	4.93E-19& 4.57E-22&	6.72\%\\
&\method(Ours)&100	&3&	\textbf{96.67\%}& \textbf{7.94E-04}&	\textbf{1.82E-18}& \textbf{11.09\%}\\
\bottomrule
\end{tabular}
\caption{Our method \method outperforms CGMH by generating sentences that satisfy more constraints, are of good quality and are likely to be natural language. Column Valid\% shows the percentage of generated sentences that satisfy all constraints, which \method clearly leads baselines. In addition, \method has better acceptance rates (Accept\%). The language generated by \method is also of good quality, because it matches other models in language model
scores $P_{\mathrm{GPT-2}}(x)$. Multiplying both the language model score and the constraint score, the sentences generated by \method tend to attain higher stationary probability $\pi(x)$. %For interrogative sentence generation with keywords, our method can generate sentences that are similar to ground-truth questions (measured by the BLEU score).
}
\label{table-results}
\end{table*}

%% file: 5.2.table2.tex
\begin{table}[!ht]
\small
\centering
\begin{tabular}{l|l}
\toprule
% Input & waste heat water \\
Keys & waste heat water \\
CGMH & what \pink{waste} is there, it seems now? \\
\method & where was the \pink{waste - water heater}? \\
\midrule
% Input & responses protect lungs mechanically \\ 
% & ejecting pathogens respiratory system \\
Keys & responses protect lungs \\
CGMH & how can immune \pink{responses} also occur by  \\ &not only infecting pathogens in the\\ &  central nervous system? \\
\method & what \pink{responses} do your \pink{lungs} have to \pink{protect} 
\\ & you from pathogenic bacteria? \\
\midrule
% Input & median temperature winter \\
Keys & median temperature winter \\
CGMH & what do you mean we have \pink{median temperature} \\ & \pink{winter} and spring, anyways?\\ 
\method & what is the \pink{median temperature} range in the \\ & \pink{winter} months? \\
\midrule
% Input & year university warsaw established \\
% Keys & university warsaw established \\
% \method & when was the technical \pink{university} of \pink{warsaw} \\ & first formally \pink{established}? \\
% CGMH & polish polytechnical institute - \pink{university} of \\ & technology \pink{warsaw} - was \pink{established} here \\ & in 1964? \\
% \midrule
% Input&french catholics concentrated france\\
Keys &catholics concentrated france \\%\midrule
CGMH & the \pink{catholics} are now mainly concentrated there. \\
\method &why are the french roman \pink{catholics} so densely \\& \pink{concentrated} in southern \pink{france}?\\
\bottomrule
% \multicolumn{2}{c}{Interrogative Sentences} \\ 
% \multicolumn{2}{c}{} \\
% \toprule
% Keys & seat \\ 
% \method & please get up from your \pink{seat} \\
% CGMH& go on in and take your \pink{seat}  \\ \midrule

% Keys & careful \\ 
% \method  & please be so very very \pink{careful}.     \\ 
% CGMH& and please be a very very \pink{careful} \\ \midrule

% Keys &turn, lights \\
% \method & \pink{turn} on the lights all the time \\
% CGMH & \pink{turn} on near all the main \pink{lights}\\ \midrule

% Keys &close, window \\
% \method& stay \pink{close} enough to the \pink{window}     \\
% CGMH& stick \pink{close} enough to meet the \pink{window}     \\ \midrule				
% Keys& nice, weekend \\
% \method&have yourself a very \pink{nice} private \pink{weekend}\\
% CGMH& please be \pink{nice} about spending the \pink{weekend}\\ 
% \bottomrule
% \multicolumn{2}{c}{(2) Imperative Sentences} \\
\end{tabular}
\caption{Case study of generating interrogative sentences with keywords, where Keys stands for keywords. Full case study is in the supplementary materials.}\label{tab:case}
% \vspace{-0.7cm}
\end{table}

%Our methods can generate sentences which satisfy the grammar rules, while the output from baseline lack completeness.

%% file: 5.3.table3.tex
\begin{table}[!htb]
\centering
\small
\begin{tabular}{l|rrrrr}
\toprule
Tasks             & Method & $\pi(x)$ & $P_{\text{GPT-2}}$     & Accept\%  & Senti \\ 
% \midrule
% All               & Ours   & \textbf{7E-04}              & \textbf{1E-18} & \textbf{11.09\%} & \textbf{0.7028}    \\ 
%                   & CGMH   & 4E-19              & 4E-22 & 6.72\%  & 0.6059    \\ 
\midrule
\multirow{ 2}{*}{P2P} & CGMH   & 9E-19              & 8E-22 & 8.16\%  & 0.8647    \\ 
& \method   & \textbf{4E-04}              & \textbf{2E-18} & \textbf{12.23\%} & \textbf{0.8801}    \\ 
                  
\midrule
\multirow{ 2}{*}{N2P} & CGMH   & 5E-20              & 6E-23 & 5.65\%  & 0.3470    \\
& \method   & \textbf{1E-03}              & \textbf{7E-19} & \textbf{9.91\%}  & \textbf{0.5254}    \\ 
                  
\bottomrule
\end{tabular} 
\caption{Generate sentences with positive sentiment. Half of the input are extracted from positive sentences (P2P), and the other half are from negative (N2P), which are harder to transform to positive sentences. 
% For the harder part (N2P), out method outperforms baseline more significantly. ``Sentiment" means the score evaluated by the pre-trained sentiment analysis model.
}
\label{table-sentiment}
\end{table}

%% file: 5.4.table4.tex
\begin{table}[!htb]
\begin{tabular}{l|rrr}
\toprule
Methods & $\pi(x)$ & $P_{\text{GPT-2}}(x)$ & Sentiment \\ \midrule
CtrlGen  & 3.19E-07 & 4.64E-22 & 0.4614     \\
\method & \textbf{1.16E-03}  & \textbf{7.07E-19}    & \textbf{0.5254}\\ 
\bottomrule
\end{tabular}
\caption{Compare with CtrlGen~\citep{DBLP:conf/icml/HuYLSX17} over the N2P subtask with acceptance rate, language score and sentiment score metrics.}
\label{tab:ctrl-gen}
\end{table}

%% file: 6-conclude.tex
\section{Conclusions}
We propose a framework for constraint-driven language generation via sampling and combinatorial constraint satisfaction. 
Our solution strategy is to sample sentences from the constrained space with probability proportional to the scores of the language model. 
To better handle the combinatorial constraints, a tree search is embedded into the proposal process of MCMC to suggest candidate proposals that satisfy more constraints.
%
%We apply our  approach in three applications.
%
Experiments demonstrate that our approach generates sentences that satisfy more constraints, are of good quality and are likely to be close in quality to the natural language.

%% file: 7-supp.tex
\appendix
\section{Appendix}
\subsection{Detailed Experiment Settings}
\label{sup-exp-settting}
In this section, we detail our experimental settings for interrogative, imperative, and sentimental sentence generation tasks, along with the process of human evaluation.

% The main step of every sampling method is to select a sentence with the highest $\pi(x)$ value among all the generated sentences. 
In the expression of stationary distribution Eq.\eqref{eq:pi}, the first term $P_{\mathrm{LM}}(x)$ is evaluated by the BERT model, which is based on the huggingface's BERT implementation~\cite{Wolf2019HuggingFacesTS}. We use BERT-base in our experiments, with hyper-parameters: L=12, H=768, A=12, Total Parameters=110M. To evaluate the term $P_{\mathrm{LM}}(x)$ with BERT model, we multiply the BERT score of masking and querying the conditional probability of each word in sentence $x$, close in form of the pseudo-likelihood \cite{wolfinger1993generalized}. Since we only requires $\pi(x)$ to be proportional to $P_{\mathrm{LM}}(x)$ times the constraint score, $P_{\mathrm{LM}}(x)$ does not need to be normalized.
%
%We use huggingface's BERT implementation~\cite{Wolf2019HuggingFacesTS}. We use BERT-base in our experiments, with hyper-parameters: L=12, H=768, A=12, Total Parameters=110M.
% 
%Our computational infrastructure is a cluster composed of Dell compute nodes with Intel Xeon processors and Nvidia Tesla GPUs.

%with constraints and input partial sentences, 

% 
% $P_{\mathrm{LM}}(x)$ within $\pi(x)$ is evaluated by multiplying the BERT score of masking and querying the conditional probability of each word in sentence $x$. This score is close in the form of the pseudo-likelihood \citep{wolfinger1993generalized} and measures the goodness of a sentence. 
% Since we only requires $\pi(x)$ to be proportional to $P_{\mathrm{LM}}(x)$ times the constraint score, $P_{\mathrm{LM}}(x)$ does not need to be normalized.
%
% We use huggingface's BERT implementation~\citep{Wolf2019HuggingFacesTS}.
% Since the $\pi(x)$ does not require normalization, this can be achieved by setting [mask].... 
% The temperate hyper-parameter is set to $\beta=10^{-10}$.
% We define different constraints for different tasks. 
% The detailed constraints and input can be seen in section \ref{exp-ques} and \ref{exp-imp}. 

\subsubsection{Interrogative Sentences Generation}
According to the adapted definition of interrogative sentence grammar, the first word should be a question word, and there should be an auxiliary verb at a suitable position. The constraint definition for interrogative sentences is in section \ref{qg-constraints}. In our actual implementation, we also enforce that there should be only one question word and one auxiliary verb in the sentence in order to improve the quality of generated sentences. The question words include \textit{what, when, where, which, who, whom,  whose,  why,  how}; the auxiliary verbs include \textit{do, does, did, be, am, are, is, was, were, shall, will, should, would, can, could, may, might, must}.
%
% We conduct experiments on interrogative sentence generation with and without additional keyword constraints.
% 

For the task of generating interrogative sentences with keywords, we also enforce the keyword only appear once in the sentence. 

The dataset of this task is based on the SQuAD 2.0 dataset~\citep{DBLP:conf/acl/RajpurkarJL18}, where we select 600 questions and removing the stop words using the Rake toolkit~\citep{rose2010automatic}.

%The input keywords are generated by removing the stopping words from 600 selected questions in SQuAD 2.0~\citep{DBLP:conf/acl/RajpurkarJL18} using the Rake toolkit~\citep{rose2010automatic}. 

\subsubsection{Imperative Sentences Generation}
The dataset for generating imperative sentences is retrieved from\footnote{https://github.com/lettergram/sentence-classification}. We select 300 sentences and extract the keywords from the sentences as our input.
According to the grammar of imperative sentences, we need to verify if the word is a present tense verb. In the implementation, we use the POS tag information in WordNet and Stanford CoreNLP as the criterion for deciding the word POS tag of the given word. We first select all the words with at least one verb meaning in WordNet~\citep{DBLP:journals/cacm/Miller95}, then use Stanford CoreNLP~\citep{manning-EtAl:2014:P14-5} to get POS tags for each word and only preserve the present tense form of verbs. %filter out variant forms of verbs (like past tense).
% (VBD), present participle(VBG), past participle(VBN) and 3rd person sing. present(VBZ) forms. Then we get all base-formed verbs in the vocabulary. 

\subsubsection{Sentiment Sentence Generation}
This application requires the set of input keywords and an external sentiment classifier, which is used to estimate whether the sentiment of the sentence is positive or not. 
To estimate the sentiment score of the sentences, we train a sentiment analysis model with fastText~\citep{joulin2017bag} on Yelp Review Polarity dataset~\citep{zhang2015character}. 
The input keywords are extracted from 300 selected sentences in the Yelp test set. Half of the original sentences are positive, and the other half are negative (which is harder to transform to positive sentences).

With input keywords of positive and negative sentiment, we enforce the model to generate sentences with positive sentiment. The second sub-task with negative sentiment keywords is much more difficult than the sub-task with positive sentiment keywords, as it requires transforming from negative to positive sentiment.

% The baseline method for comparison is CtrlGen.

% \subsection{Human Evaluation}
% \label{sup-human-eval}
% We conduct human evaluation for  interrogative sentences generated with keywords. 
% %
% We present human participants from the Amazon Mechanical Turk with a pair
% of sentences at a time. One sentence is generated by our \method model and the other one is from the baseline CGMH. 
% %
% We ask human participants which sentence is better in terms of fluency and grammar. 
% %
% % In this evaluation, $32.72\%$ of the participants agree that our \method model generates better sentences. 
% %
% In terms of the experimental setting, we use 100 sentence pairs generated by CGMH and \method with the same keyword inputs. We randomly split these 100 test sentence pairs into 5 survey groups, and then deploy them on the Amazon Mechanical Turk.
% We randomly assign human participants to survey groups.
% %for native speakers to vote which one is better. 
% When showing the sentence pairs, we also provide the keywords that the sentences 
% must contain. 
% %
% We ask human participants to vote which sentence in the pair is better. 
% %
% We use a gold-standard question to detect if the voter is randomly doing the survey. 
% %
% Every valid survey contains a randomized set of 20 questions. We received in all 580 votes. Each question pair receives votes ranging from $5$ to $11$. As shown in Table~\ref{tab:human}, sentences from our model receive almost twice times of votes than the baseline, which suggests that the sentences generated by our approach is better in human evaluation. 

\subsection{Case Studies}
\input{7.1.table.case}
\label{sup-case-study}
As shown in Table~\ref{tab:case-full}, we compare some output sentences of our method with the baseline using the same inputs and keywords.
% for both interrogative and imperative sentence generation tasks. 
From these cases, we can see that the baseline sometimes generates awkward or disordered sentences.
For example, the baseline generates one sentence:``\textit{how was lower normandy ever truly founded?}''. Although this sentence seems to satisfy the constraints of an interrogative sentence, its meaning is awkward. The sentence generated by our method is ``\textit{when was the duchy of normandy founded?}'', which is more realistic. Also, the sentence from the baseline  ``\textit{and please be a very very careful}'' does not follow imperative grammar, and ``\textit{the catholics are now mainly concentrated there}'' is not a question.

%% file: 7.1.table.case.tex
\begin{table}[!ht]
\small
\centering
\begin{tabular}{l|l}
\toprule
% Input & waste heat water \\
% Keys & waste heat water \\
% \method & where was the \pink{waste - water heater}? \\
% CGMH & what \pink{waste} is there, it seems now? \\
% \midrule
% % Input & responses protect lungs mechanically \\ 
% % & ejecting pathogens respiratory system \\
% Keys & responses protect lungs \\
% \method & what \pink{responses} do your \pink{lungs} have to \pink{protect} 
% \\ & you from pathogenic bacteria? \\
% CGMH & how can immune \pink{responses} also occur \\ & by not only infecting pathogens in \\ & the central nervous system? \\
% \midrule
% % Input & median temperature winter \\
% Keys & median temperature winter \\
% \method & what is the \pink{median temperature} range in the \\ & \pink{winter} months? \\
% CGMH & what do you mean we have \pink{median temperature} \\ & \pink{winter} and spring, anyways?\\ 
% \midrule
% % Input & year university warsaw established \\
Keys & university warsaw established \\
\method & when was the technical \pink{university} of \pink{warsaw} \\ & first formally \pink{established}? \\
CGMH & polish polytechnical institute - \pink{university} of \\ & technology \pink{warsaw} - was \pink{established} here \\ & in 1964? \\
\midrule
Keys & organization charge running\\
\method & who would \pink{charge} her with \pink{running} such an \\ & \pink{organization}? \\
CGMH & who else would \pink{charge} him with \pink{running} a  \\ &very profitable business? \\
\midrule
Keys & tribes khan fight \\
\method &  what \pink{tribes} would \pink{fight} back against the \\ &genghis  \pink{khans}?\\
CGMH & why else would tribesmen like gen. and gen. \\& genghis \pink{khan} \pink{fight} them off? \\
\midrule
Keys & european travel amazon \\
\method & why did early \pink{european} explorers not \pink{travel} to \\& amazonia? \\
CGMH & see below, also : did any \pink{european} settlers ever \\& \pink{travel} to build the " first north american sailing \\& canoes "? \\
\midrule
Keys & economic growth schooling \\
\method & how do \pink{economic} \pink{growth} rates in the united \\& states make children receive high - quality \\& \pink{schooling}? \\
CGMH & what good is \pink{economic growth} in comparison \\& with being among the best in public \pink{schooling}? \\
% \midrule
% % Input&french catholics concentrated france\\
% Keys &catholics concentrated france \\%\midrule
% \method &why are the french roman \pink{catholics} so densely \\& \pink{concentrated} in southern \pink{france}?\\
% CGMH & the \pink{catholics} are now mainly concentrated there. \\
\bottomrule
\multicolumn{2}{c}{(1) Interrogative Sentences} \\ 
\multicolumn{2}{c}{} \\
\toprule
Keys & seat \\ 
\method & please get up from your \pink{seat} \\
CGMH& go on in and take your \pink{seat}  \\ \midrule

Keys & careful \\ 
\method  & please be so very very \pink{careful}.     \\ 
CGMH& and please be a very very \pink{careful} \\ \midrule

Keys &turn, lights \\
\method & \pink{turn} on the lights all the time \\
CGMH & \pink{turn} on near all the main \pink{lights}\\ \midrule

Keys &close, window \\
\method& stay \pink{close} enough to the \pink{window}     \\
CGMH& stick \pink{close} enough to meet the \pink{window}     \\ \midrule				
Keys& nice, weekend \\
\method&have yourself a very \pink{nice} private \pink{weekend}\\
CGMH& please be \pink{nice} about spending the \pink{weekend}\\ 
\bottomrule
\multicolumn{2}{c}{(2) Imperative Sentences} \\
\end{tabular}
\caption{Case study of generating interrogative and imperative sentences with keywords, where Keys stands for keywords.}\label{tab:case-full}
\vspace{-0.7cm}
\end{table}

%Our methods can generate sentences which satisfy the grammar rules, while the output from baseline lack completeness.